\pdfoutput=1
\documentclass[11pt]{article}

\usepackage{emnlp2021}

\usepackage{times}
\usepackage{latexsym}
\usepackage{url}
\usepackage{paralist}
\usepackage{multirow}
\usepackage{verbatim}
\usepackage{amsfonts}
\usepackage{amssymb}
\usepackage{amsmath}
\usepackage{algorithm}
\usepackage{algpseudocode}
\usepackage{graphics}
\usepackage{epsfig}
\usepackage{multirow}
\usepackage{makecell}
\usepackage{booktabs}

\usepackage[T1]{fontenc}
\usepackage[utf8]{inputenc}

\usepackage{microtype}

\newcommand\blfootnote[1]{%
  \begingroup
  \renewcommand\thefootnote{}\footnote{#1}%
  \addtocounter{footnote}{-1}%
  \endgroup
}

\DeclareUnicodeCharacter{202F}{\,}

\title{HUMBO: Bridging Response Generation and Facial Expression Synthesis}

\author{Shang-Yu Su$^\star$\quad Po-Wei Lin$^\star$\quad Yun-Nung Chen \\
Department of Computer Science and Information Engineering\\
National Taiwan University\\
\texttt{\{f05921117, r09922a24\}@csie.ntu.edu.tw\quad y.v.chen@ieee.org}
}

\begin{document}
\maketitle
\begin{abstract}
Spoken dialogue systems that assist users to solve complex tasks such as movie ticket booking have become an emerging research topic in artificial intelligence and natural language processing areas. 
With a well-designed dialogue system as an intelligent personal assistant, people can accomplish certain tasks more easily via natural language interactions. 
% Today there are several virtual intelligent assistants in the market; however, most systems only focus on single modality, such as textual or vocal interaction.
Today there are several virtual intelligent assistants in the market; however, most systems only focus on textual or vocal interaction.
% A multimodal interface has various advantages:
% (1) allowing human to communicate with machines in a natural and concise form using the mixture of modalities that most precisely convey the intention to satisfy communication needs, and
% (2) providing more engaging experience by natural and human-like feedback.
% This paper explores a brand new research direction, which aims at bridging dialogue generation and facial expression synthesis for better multimodal interaction.
% The goal is to generate dialogue responses and simultaneously synthesize corresponding visual expressions on faces, which is also an ultimate step toward more human-like virtual assistants.
In this paper, we present \textbf{HUMBO}, a system aiming at generating dialogue responses and simultaneously synthesize corresponding visual expressions on faces for better multimodal interaction.
HUMBO can 
(1) let users determine the appearances of virtual assistants by a single image,
and (2) generate coherent emotional utterances and facial expressions on the user-provided image.
This is not only a brand new research direction but more importantly, an ultimate step toward more human-like virtual assistants.
% This paper explores a brand new research direction, aiming at generating dialogue responses and simultaneously synthesize corresponding visual expressions on faces, which is also an ultimate step toward more human-like virtual assistants. 
\blfootnote{$^\star$The first two authors contributed to this work equally.}
\end{abstract}

\section{Introduction}
% Chatbots and virtual assistants are also ripened fruits of the recent progress of AI.
The recent advance of deep learning has inspired many applications of neural dialogue systems~\cite{wen2017network,bordes2017learning}.
A typical dialogue system pipeline can be divided into several components: a speech recognizer that transcribes a user's speech input into texts, a natural language understanding module (NLU) to classify the domain along with domain-specific intents and fill in a set of slots to form a semantic frame~\cite{hakkani2016multi}. 
Following a dialogue state tracking (DST) module that predicts the current dialogue state according to the multi-turn conversations, then the dialogue policy determines the system action for the next step given the current dialogue state~\cite{su2018discriminative}.
% Finally the semantic frame of the system action is then fed into a natural language generation (NLG) module to construct a response utterance to the user~\cite{wen2015semantically, su2018natural, su2018investigating}. 
Finally the semantic frame of the system action is then fed into a natural language generation (NLG) module to construct a response utterance to the user.

\begin{figure}[t!]
\centerline{\includegraphics[width=\linewidth]{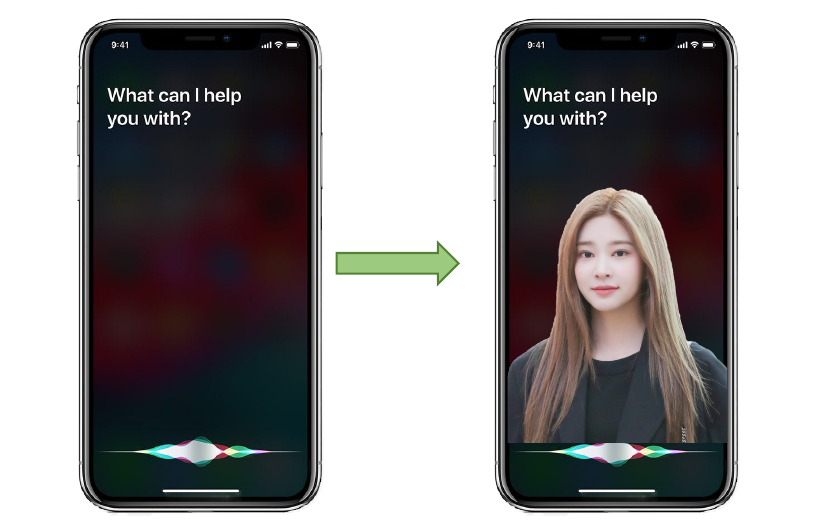}}
\vspace{-2mm}
	\caption{The illustration of the idea of characterization of virtual assistants, the virtual assistants have human-like appearance. In this work, we aim to endure users with the ability of determining the appearance of their virtual assistants.}
	\label{fig:characterization}
\end{figure}

% multimodal vqa image caption -> dialogue dataset 
Nowadays, several virtual intelligent assistants show up in the market, such as Apple Siri, Google Assistant, Microsoft Cortana, and Amazon Alexa. 
However, most of these systems only focus on single or monotonous modality, such as textual or vocal interaction.
Although the existing systems have shown capability of enabling users to perform basic information inquiries and helping simple daily activities, real human-human conversation actually involve multiple modalities of information.
Communication between humans is complex and often requiring mixture of expression mode to easily precisely exchange information, for example, using both gestures and voice. 
Multimodal dialogues reflect more human-like behavior, however, this research topic of is barely explored due to its difficulty in data collection, cross-modality reasoning, and various aspects.
In this paper, we explores a brand new research direction, which aim to bridge response generation and facial expression synthesis.
% The goal is to generate dialogue responses and simultaneously synthesize corresponding faces, which is also an ultimate step toward more human-like virtual assistants. 
The idea is inspired by talking head tasks \cite{suwajanakorn2017synthesizing}, the line of research in synthesizing talking faces from speech, in which the face is animated to mimic the continuous time-varying context (i.e. talking) and affective states carried in the speech.

The proposed task is to generate natural language utterances and then construct realistic faces based on the generated sequences.
In other words, the synthesized facial expression should be relevant to certain semantic concept in the generated sentences, in this paper, we model the shared semantics of the two modalities by \textbf{emotion}.
Since emotions are expressed through a combination of verbal and non-verbal channels, like gestures, facial expression, speech, and spoken content, therefore it could be viewed as \emph{the intersection and the bridge of various modes}. 
% In spoken language, emotion detection has been a widely explored field, there are two types of analysis available to detect emotion: sentiment analysis and emotion analysis.
% In sentiment analysis \cite{socher2013recursive}, the goal is to detect sentiment from the given user input text, which is generally a bipolar or tri-polar (positive, negative and neutral) feeling, while in emotion analysis \cite{chen2018emotionlines} we can detect types of generic feelings such as happy, sad, anger, disgust, fear and surprise from the given input.  

% In this work, we set out to bridge dialogue generation and facial expression synthesis, by combining the two functions, the dialogue agent could automatically generate utterances along with corresponding facial expressions.
The proposed concept is a new research direction of multimodal dialogues and also an ultimate step towards more human-like virtual assistants.
We believe such characterization (Figure \ref{fig:characterization}) would be one of the important developing direction of chatbots in the future. 
% Furthermore, a dataset preparation strategy and a training framework are also introduced.
We hereby propose \textbf{HUMBO} (\textbf{HUM}an-like \textbf{BO}t), a framework consisting of two main components: 
(1) a GAN-based facial expression generator, and (2) a GPT-based multi-task language generator.
HUMBO can let users determine the appearances of virtual assistants by a single image,
and generate coherent emotional utterances and facial expressions on the user-provided image.

\begin{figure*}[t!]
\centerline{\includegraphics[width=\linewidth]{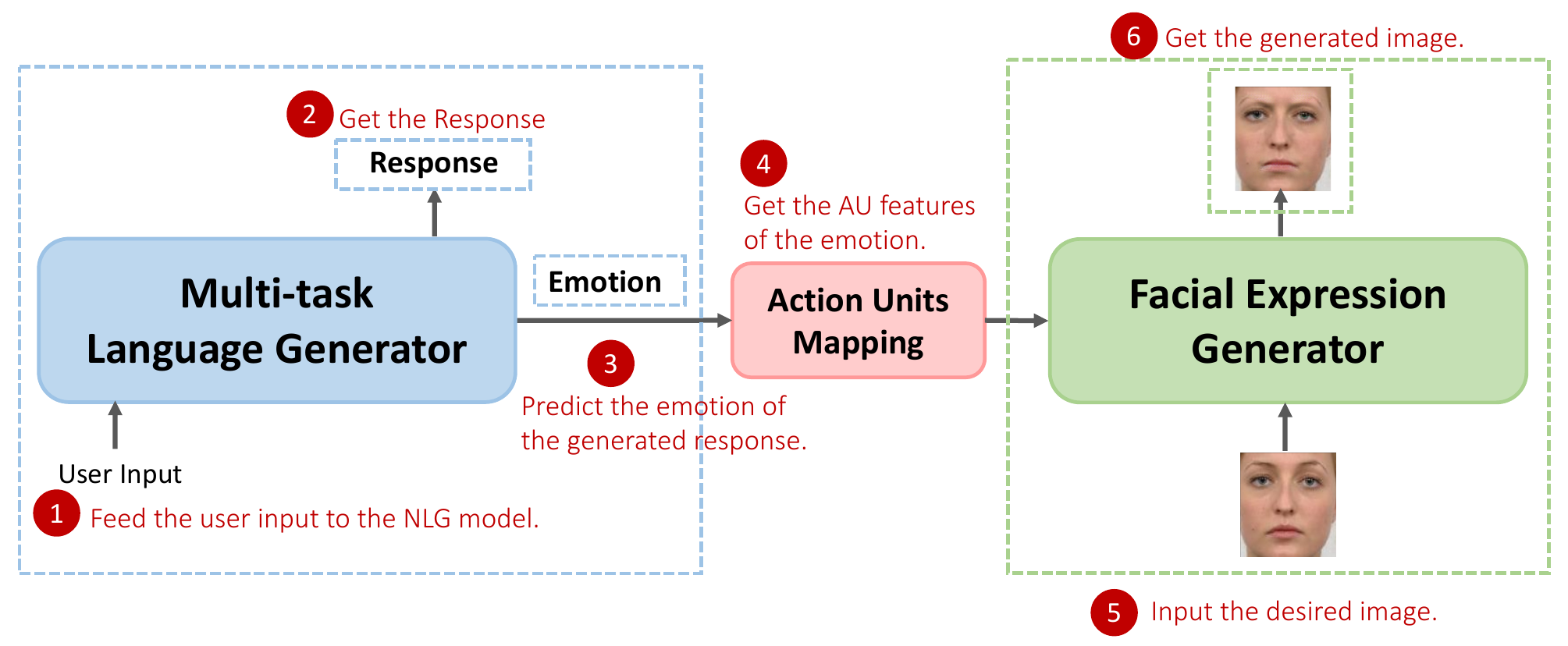}}
\vspace{-2mm}
	\caption{The model illustration of the proposed pipeline.}
	\label{fig:framework}
\end{figure*}

\section{HUMBO}
In this section, we first describe the data preparation strategy and then detail the technology we adopted in HUMBO.

% \subsection{Task Description}
% Given a dialogue consisting of $n$ sentence-level features $d = \{ x_{i} \}_{i=1}^{n}$ sampled from a set of dialogues $\mathcal{D}$.
% The core goal of response generation is to generate proper next responses based on preceding dialogue contexts.
% A typical strategy for the optimization problem is based on maximum likelihood estimation (MLE) of the parameterized conditional distribution by the learnable parameters $\theta$ formulated as below:

% \begin{equation*}
%     \theta^{*} = \underset{\theta}{\mathrm{argmax}} P(x_{i} \mid x_{1,2,...i-1} ; \theta ).
% \end{equation*}

% % where $f$ is the learned mapping function.
% The proposed task is to generate natural language utterances, predict the emotion information of the spoken content, and then construct realistic faces based on the emotion distribution.
% Because of such multimodal scenario, the sentence-level features $x$ contains not only utterances but emotion information and corresponding facial expression, $x=(s,e,f)$.
% In other words, the goal of the generative model is to estimate joint probability of spoken content $s$, emotion distribution $e$, and facial expression $f$.
% The MLE objective is hereby reformulated as:
% \begin{align*}
% \theta^{*} =\underset{\theta}{\mathrm{argmax}} P((s, e, f)_{i} \mid (s, e, f)_{1,2,...i-1} ; \theta ).
% \end{align*}

\subsection{Data Preparation}
In this work, we focus on response generation, however, most of the textual emotional datasets consist of emotion labels of only individual words, sentences or documents, which makes it challenging to discuss the contextual flow of emotions.
Although the IEMOCAP database \cite{busso2008iemocap} provides emotion labels for each utterance, it was created by actors performing emotions, and hence carries the risk of overacting \cite{chen2018emotionlines}. 
Moreover, the annotators label the emotions by watching the videos instead of reading the transcripts , which means the annotators may make the decision only depend on the facial  expression or the prosodic features without realizing the meaning of the words.
Considering such potential bias, we decide to combine separate vision and language datasets according to our needs.  

For language part, we choose the Emotion Detection dataset from Emory NLP \cite{zahiri2017emotion}, which is collected from the transcripts of the TV show \emph{Friends}.
The corpus comprises 97 episodes, 897 scenes, and 12,606 utterances, where each utterance is annotated with one emotion label.
The emotions have 7 types in total, the six primary emotions in the Feeling Wheel \cite{Willcox1982TheFW}, sad, mad, scared, powerful, peaceful, joyful, and a default emotion of neutral.
Each label was accomplished by 4 crowd workers, and for each utterance in a label, the emotion with the highest number of votes was set as the gold label of the utterance.

% For language part, we choose EmotionLines \cite{chen2018emotionlines}, which contains a total of 29245 labeled utterances from 2000 dialogues.
% Each utterance in dialogues is labeled with one of seven emotions, six Ekman’s basic emotions plus the neutral emotion.
% Each label was accomplished by 5 workers, and for each utterance in a label, the emotion with the highest number of votes was set as the gold label of the utterance.
% Those utterances with more than two different emotions voted were put into the non-neutral category.
% Therefore the dataset has a total of 8 types of emotion labels, anger, disgust, fear, happiness, sadness, surprise, neutral, and non-neutral.  

On the other hand, Radboud Faces Database (RaFD) \cite{langner2010presentation} has 8 binary labels for facial expressions, namely sad, neutral, angry, contemptuous, disgusted, surprised, fearful and happy.
In total, the set contains 67 models: 20 Caucasian male adults, 19 Caucasian female adults, 4 Caucasian male children, 6 Caucasian female children, and 18 Moroccan male adults.
All models in the dataset show the above eight facial expressions with three gaze directions, photographed simultaneously from five different camera angles.
The photos were taken in a highly controlled  environment.  All displayed facial expressions were based on prototypes from Facial Action Coding System (FACS) \cite{ekman1976measuring}.
FACS was developed for describing facial expressions in terms of the so-called Action Units (AUs), which are anatomically related to the contractions of specific facial muscles.
We select frontal images, crop the head regions, and use OpenFace 2.2.0 \cite{baltrusaitis2018openface} to recognize Action Units from the images.
% Note that either the facial expression dataset or the dialogue with emotion label dataset could be chosen or collected at will.
The images in RaFD are high-quality and suitable for facial expression generation where each model was asked to make different facial expressions in a clean white background. 
However, RaFD only has 535 images in total, hence we further utilize CelebA \cite{liu2015faceattributes} in training.

\subsection{Pipeline}
\label{ssec:pipeline}
In this section, we design a pipeline composed of two main components: (1) a multi-task NLG model based on DialoGPT ~\cite{Zhang_2020} and (2) a GAN-based model for facial expression generation \cite{pumarola2018ganimation}.

\begin{figure}[t!]
\centerline{\includegraphics[width=\linewidth]{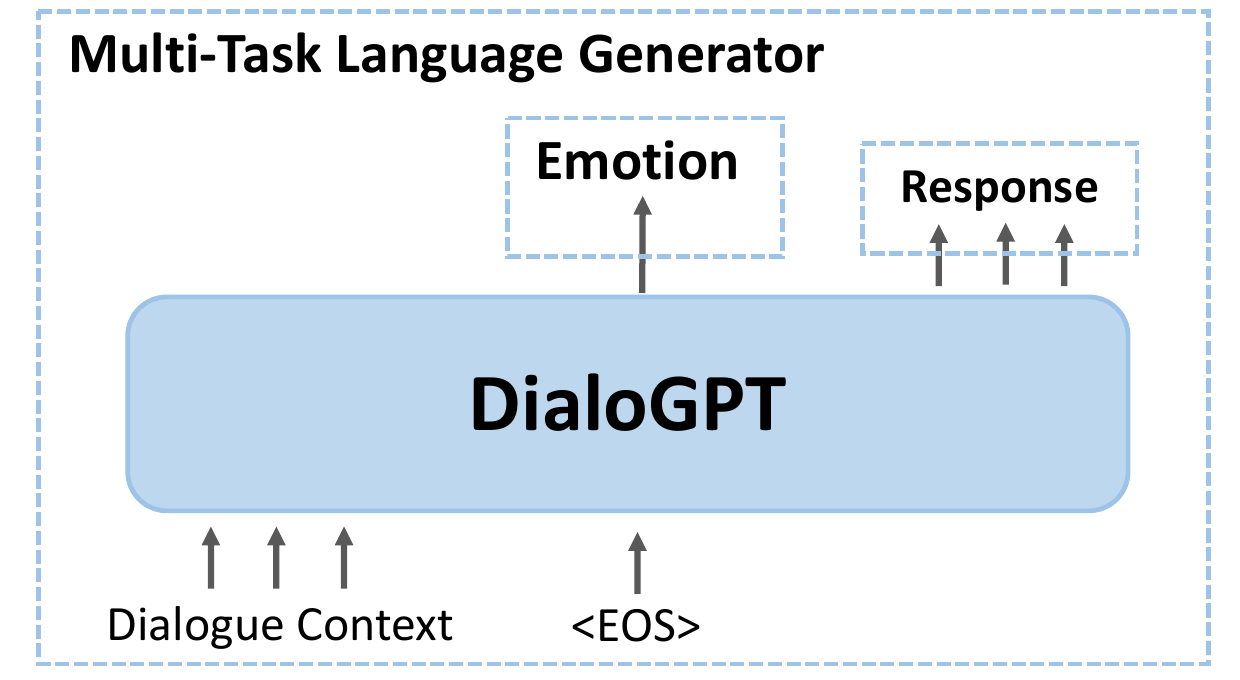}}
\vspace{-2mm}
	\caption{The architecture of the multi-task language generator, which is based on DialoGPT.}
	\label{fig:nlg}
\end{figure}

\subsubsection{Multi-task Language Generator}
The framework of the proposed NLG model is illustrated in Figure \ref{fig:nlg}, where the model architecture is based on based on DialoGPT \cite{Zhang_2020}.
DialoGPT models are trained on the basis of the GPT-2 \cite{radford2019language} architecture, they are both auto-regressive language model and uses the multi-layer transformer as model architecture.
The difference between two models is that DialoGPT focuses on dialogues where is trained on large-scale dialogue sessions extracted from Reddit discussion threads. 
Therefore we choose DialoGPT as our dialogue response generator and finetune it on the language dataset described in the previous section.
The emotion prediction is also a classification problem, where we aim to use the predicted emotion signal to bridge the language and facial expression.
Given a dialogue pair $\{(u_{t}, e_{t}), (u_{t+1}, e_{t+1}) \}$, each utterance $u_{(.)}$ has its corresponding emotion label $e_{(.)}$ and dialogue context $D$.
We utilize DialoGPT to build a multi-task language generator,
\begin{equation*}
    u_{t+1}, e_{t} = \mathrm{DialoGPT}(D),
\end{equation*}
where the goal of the generator is to generate the next utterance and identify the emotion of the current turn.
As depicted in Figure \ref{fig:nlg}, the end-of-sentence token <EOS> will be appended to the dialogue context before feeding into the model, the encoded feature at the position of <EOS>.
The predictor is a simple full-connected layer.
The multi-task loss is the combination of the language generation loss and the emotion prediction loss.
\begin{equation*}
     \mathcal{L}_\text{MLG} = \lambda_\text{u} \cdot \mathcal{L}_\text{u} + \lambda_\text{e} \cdot  \mathcal{L}_\text{e}, 
\end{equation*}
where $\lambda_\text{u}$ and $\lambda_\text{e}$ are the weight for the language generation loss and the emotion prediction loss respectively.
When doing inference, the emotion of next turn $e_{t+1}$ is what we want, hence the next  utterance $u_{t+1}$ will be fed into the model again for identifying the emotion label.

\subsubsection{Action Units Mapping}
To bridge the two models, we further design a tabular mapping from predicted emotion signal to the target attribute, which is activation of Action Units.
%, based on the anatomical information in RaFD paper \cite{langner2010presentation}.
% The tabular is built by sampling a photo in RaFD \cite{langner2010presentation} with specific emotion and using OpenFace to recognize action units. 
Instead of utilizing the deterministic anatomical information in RaFD paper \cite{langner2010presentation}, like happiness would trigger the action unit 6 and 12, we build the tabular by sampling a photo with specific emotion in RaFD and using OpenFace to recognize action units.
Since the process of recognizing AUs depends on two trained models: (1) a face detection toolkit\footnote{\url{https://pypi.org/project/face-recognition/}} and (2) OpenFace for action unit recognition, we observe that deterministic anatomical information of AU activation is not reliable. 
In contrast, extracting the AU information from the dataset we are using would be a better choice.
With the mapping, we can extract corresponding AU features for predicted specific emotion to pass into the facial expression generator.

\begin{figure}[t!]
\centerline{\includegraphics[width=\linewidth]{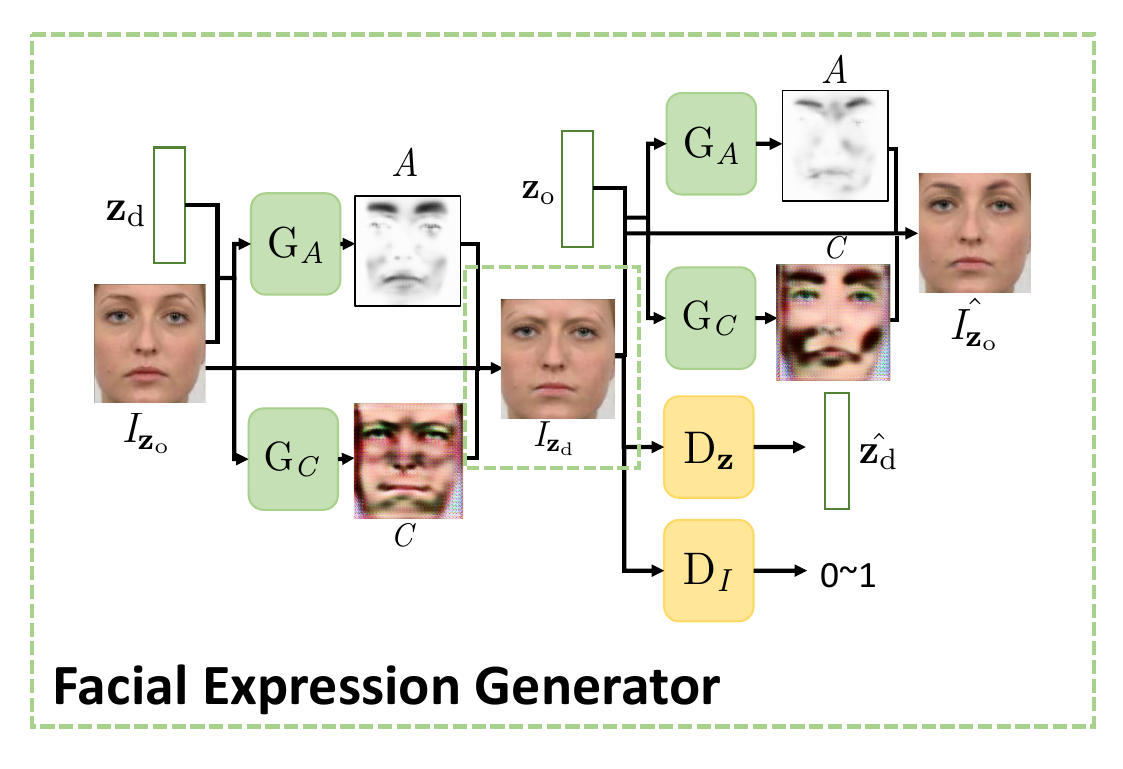}}
\vspace{-2mm}
	\caption{The architecture of the facial expression generator. The parameters in generators ($\mathrm{G}_\mathit{A}$ and $\mathrm{G}_\mathit{C}$) are shared except the last layer, while the ones in the discriminators are shared in the same way as well.}
	\label{fig:FEG}
\end{figure}

\subsubsection{Facial Expression Generator}
We utilize the GAN-based model \cite{pumarola2018ganimation} as our face generator.
As illustrated in Figure \ref{fig:FEG}, the face generator is composed of two main modules: 
(1) a pair of generators ($\mathrm{G}_\mathit{A}$ and $\mathrm{G}_\mathit{C}$) is trained to change the facial expression in image $\mathit{I}$ according to the given desired attributes $\mathbf{z}$ and 
(2) a pair of WGAN-GP-based discriminators ($\mathrm{D}_\mathit{I}$ and $\mathrm{D}_\mathbf{z}$) to examine the photo-realism and desired expression fulfillment of the generated images.

Since our goal is to manipulate the facial expression in the image $\mathit{I}$, the generator should focus only on those regions of the image that are relevant to constitute facial expressions and keep the rest elements of the image such as hair, glasses, hats or background untouched.
For this purpose, instead of directly regressing a full image, the generators predict two masks, a color mask $\mathit{C}$ and an attention mask $\mathit{A}$, by $\mathrm{G}_\mathit{A}$ and $\mathrm{G}_\mathit{C}$ respectively.
Rather than directly constructing a full image as in the typical GANs, the image is now obtained by the following formula:
\begin{align}
\label{eq:ganimation}
\mathit{I}_{\mathbf{z}_\text{d}} = \mathrm{G}(\mathit{I}_{\mathbf{z}_\text{o}} \mid \mathbf{z}_\text{d}) = (1 - \mathit{A}) \cdot \mathit{C} + \mathit{A} \cdot \mathit{I}_{\mathbf{z}_\text{o}}, 
\end{align}
where $\mathbf{z}_\text{d}$ and $\mathbf{z}_\text{o}$ represent the desired and original attributes of facial expressions respectively,
and the subscripts d and o denote the words \emph{desired} and \emph{original}.
Note that the desired attributes $\mathbf{z}_\text{d}$ are mapped from the predicted emotion tag as mentioned in the previous section.
Both the color mask and the attention mask are predicted based on the original input image and the given desired attributes, $\mathit{C} = \mathrm{G}_\mathit{C} (\mathit{I}_{\mathbf{z}_\text{o}} \mid \mathbf{z}_\text{d})$, $\mathit{A} = \mathrm{G}_\mathit{A} (\mathit{I}_{\mathbf{z}_\text{o}} \mid \mathbf{z}_\text{d})$.
The generated attention $\mathit{A}$ is a single-channel mask, indicating the preserved region from the original image.
On the other hand, the color mask $\mathit{C}$ is a RGB, three-channel mask, determining the actual facial movements.
By (\ref{eq:ganimation}), the model could focus on the pixels defining the facial movement and preserve the static region from the original image, which leads to sharper and more realistic synthesis.

Original GAN training utilizes Jensen-Shannon (JS) divergence as the loss function, which aims to maximize the probability of correctly distinguishing between real and generated data.
Since the divergence is potentially not continuous hence resulting in vanishing gradients,
in order to address the issue, we use WGAN-GP \cite{gulrajani2017improved} as our adversarial framework, which replaces JS divergence with Earth Mover Distance (Wasserstein Distance) along with a gradient penalty term.
The gradient penalty term is an alternative way to enforce the Lipschitz condition, which directly constrains the gradient norm of the critic’s output with respect to its input.
The adversarial objective is hereby formulated as below:
\begin{align}
\label{eq:adversarial}
 \mathcal{L}_\text{adv} = \mathbb{E}_{\mathit{I}_{\mathbf{z}_\text{o}} \sim \mathit{P}_\text{o}} [\mathrm{D}_\mathit{I}(\mathrm{G}(\mathit{I}_{\mathbf{z}_\text{o}} \mid \mathbf{z}_\text{d})) - \mathrm{D}_\mathit{I}(\mathit{I}_{\mathbf{z}_\text{o}})] \nonumber \\
 + \lambda_\text{gp} \mathbb{E}_{\hat{\mathit{I}} \sim \mathit{P}_\text{i}}[(\lVert \nabla_{\hat{\mathit{I}}} \mathrm{D}_\mathit{I}(\hat{\mathit{I}})\rVert_{2} - 1)^2],
\end{align}
where $\lambda_\text{gp}$ is the penalty coefficient, $\mathit{P}_\text{o}$ and $\mathit{P}_\text{i}$ represent original data distribution and the distribution of sampling uniformly along straight lines between pairs of points sampled from the original data distribution.

The attention mask $\mathit{A}$ and the color mask $\mathit{C}$ are both learned by direct end-to-end training driven by the signals provided by the discriminator.
Because the discriminator would assess the photo-realism, the attention mask would tend to saturate to all 1, leading to a complete copy of the original input image.  
To circumvent the potential issue and improve smoothness of transformation, regularization $\mathcal{L}_\mathit{A}$ is performed over attention mask distribution:
\begin{align}
\label{eq:attn}
  \lambda_\text{tv}  \mathbb{E}_{\mathit{I}_{\mathbf{z}_\text{o}} \sim \mathit{P}_\text{o}} & \bigg[\sum_{i,j}[(\mathit{A}_{i+1,j}-\mathit{A}_{i,j})^{2}+  \\
  & (\mathit{A}_{i,j+1}-\mathit{A}_{i,j})^{2}] \bigg]   - \mathbb{E}_{\mathit{I}_{\mathbf{z}_\text{o}} \sim \mathit{P}_\text{o}} [\lVert \mathit{A} \rVert_{2}], \nonumber
\end{align}
where $i$ and $j$ stand for the indexes of the attention mask matrices.
% Since the attention mask 
The first term enforce the smoothness, while the second term is the standard $l_{2}$ norm penalty.

As the training scheme is based on conditional GANs, the generators should learn to synthesize realistic data and simultaneously satisfy the given attributes $\mathbf{z}$, which are activation of Action Units.
Specifically, the discriminators should identify $\mathbf{z}_\text{d}$ from generated examples and $\mathbf{z}_\text{o}$ from original data. 
Another condition loss $\mathcal{L}_\mathbf{z}$ is hereby formulated as below:
\begin{align}
\label{eq:cond}
 \mathbb{E}_{\mathit{I}_{\mathbf{z}_\text{o}} \sim \mathit{P}_\text{o}} [\lVert \mathrm{D}_\mathbf{z} (\mathrm{G}(\mathit{I}_{\mathbf{z}_\text{o}} \mid \mathbf{z}_\text{d})) - \mathbf{z}_\text{d} \rVert^{2}_{2} + \nonumber \\ 
 \lVert \mathrm{D}_\mathbf{z} (\mathit{I}_{\mathbf{z}_\text{o}}) - \mathbf{z}_\text{o} \rVert^{2}_{2}].
\end{align}

The above described objectives encourage the generators to render realistic facial expression $\mathit{I}_{\mathbf{z}_\text{d}}$ according to desired attributes $\mathbf{z}_\text{d}$. 
However, the generated face is not guaranteed to correspond to the same person in the input original image $\mathit{I}_{\mathbf{z}_\text{o}}$.
In this work, the \emph{cycle consistency loss} $\mathcal{L}_\text{cycle}$ \cite{zhu2017unpaired} is utilized to regularize the learning inclination to preserve the identity in the original input:
\begin{align}
\label{eq:cycle}
 \mathbb{E}_{\mathit{I}_{\mathbf{z}_\text{o}} \sim \mathit{P}_\text{o}} [\lVert \mathrm{G} (\mathrm{G}(\mathit{I}_{\mathbf{z}_\text{o}} \mid \mathbf{z}_\text{d}) \mid \mathbf{z}_\text{o}) - \mathbf{z}_\text{o} \rVert_{1}].
\end{align}
Markovian discriminator (PatchGAN) \cite{isola2017image} is introduced to restrict attention to the structure in local image patches to model high-frequency region, while $l_{1}$ norm it utilized to capture low-frequency structure.
Next, equations (\ref{eq:adversarial}) to (\ref{eq:cycle}) with their corresponding coefficients are combined into the full objective:
\begin{align*}
% \label{eq:full}
 \mathcal{L}_\text{FEG} = \mathcal{L}_\text{adv} + \lambda_\mathit{A} \mathcal{L}_\mathit{A} + \lambda_\mathbf{z}\mathcal{L}_\mathbf{z} + \lambda_\text{cycle} \mathcal{L}_\text{cycle},
\end{align*}
where $\lambda_\mathit{A}$, $\lambda_\mathbf{z}$, and $\lambda_\text{cycle}$ are the hyperparameters controlling the importance of each loss term.
Finally, we aim to solve the following minimax problem:
\begin{align*}
% \label{eq:minimax}
 \mathrm{G}^{*} =  \mathrm{arg} \min_\mathrm{G} \max_\mathrm{D} \ \mathcal{L}_\text{FEG}.
\end{align*}

\section{Demo and Discussion}
% \begin{figure}[t!]
% \centerline{\includegraphics[width=\linewidth]{figures/example.pdf}}
% \vspace{-2mm}
% 	\caption{The preliminary results generated by our proposed model. For each speaker, a random-sampled face with neutral emotion is assigned; the utterances and facial expression is then generated as described in the section \ref{ssec:pipeline}.}
% 	\label{fig:example}
% \end{figure}
\begin{figure}[t!]
\centerline{\includegraphics[width=\linewidth]{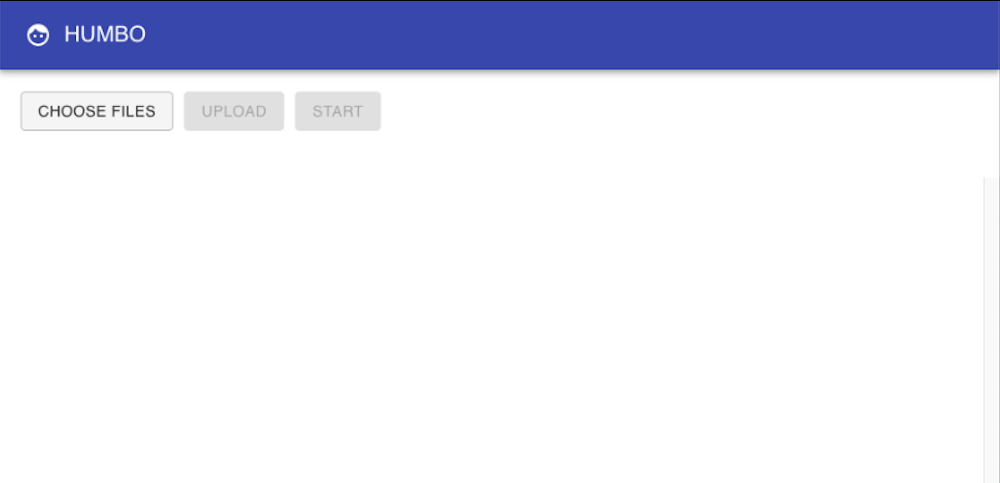}}
\vspace{-2mm}
	\caption{Before everything started, the user is required to upload a image by clicking "Choose files" to select and "Upload" to upload it.}
	\label{fig:screenshot_1}
\end{figure}

\begin{figure}[t!]
\centerline{\includegraphics[width=\linewidth]{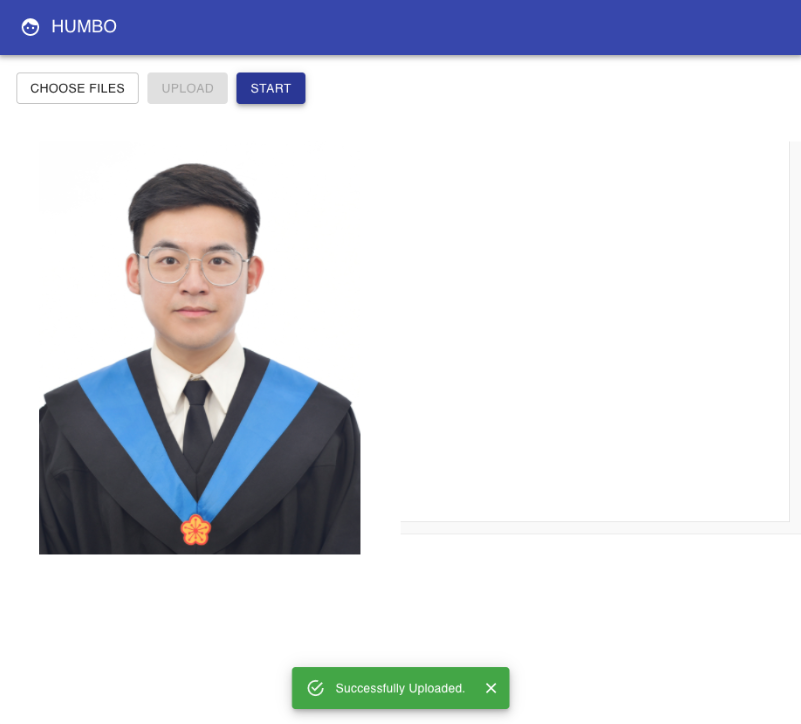}}
\vspace{-2mm}
	\caption{After successfully uploading the image, there will be a green notification popped up at the bottom, the user can then start talking with HUMBO by "Start".}
	\label{fig:screenshot_2}
\end{figure}

\begin{figure}[t!]
\centerline{\includegraphics[width=\linewidth]{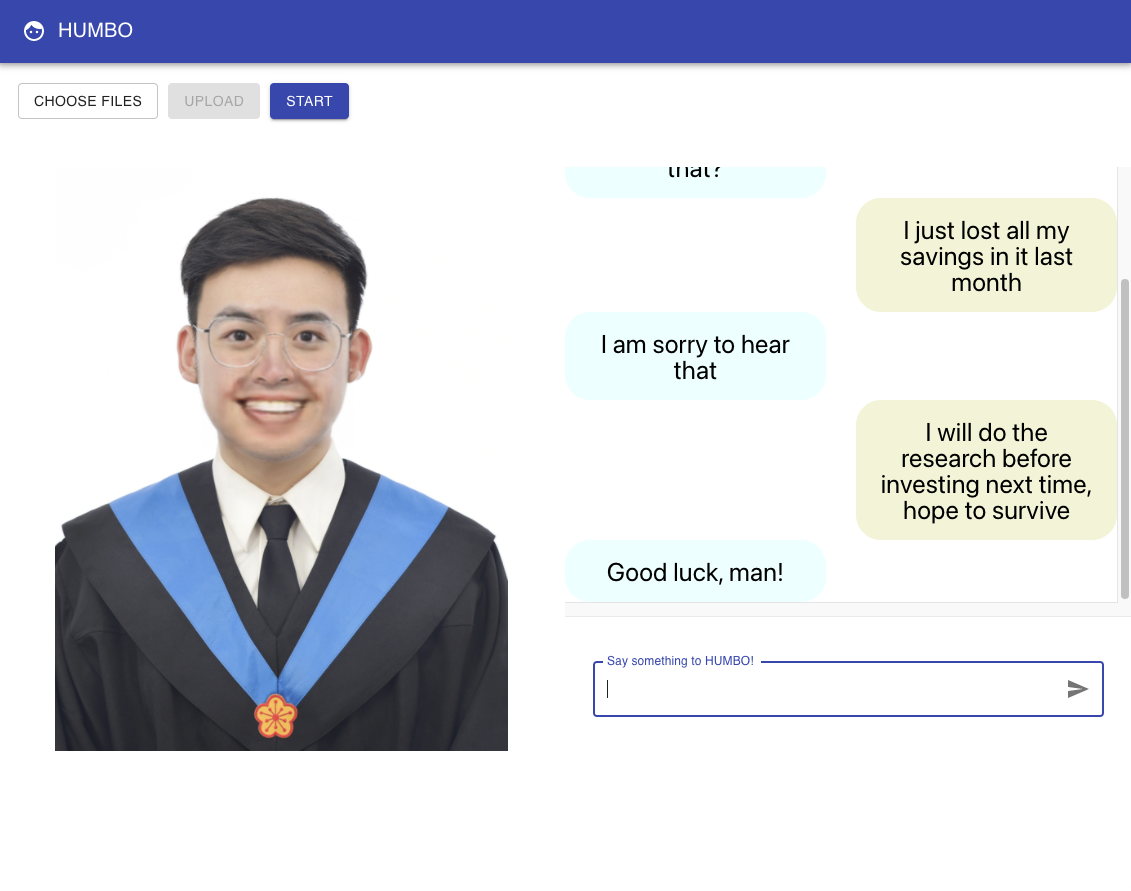}}
\vspace{-2mm}
	\caption{At each turn, the facial expression of the virtual assistant would change along with the emotion of the generated utterance.}
	\label{fig:screenshot_3}
\end{figure}

HUMBO is a web-based system: 
(1) the front-end is built by the JavaScript framework ReactJS\footnote{\url{https://reactjs.org/}}, 
and (2) the back-end is built by Python library Flask.
The usage flow of HUMBO is designed to be very intuitive:
(1) first, select the image and upload it to determine the appearance of your virtual assistant (Figure \ref{fig:screenshot_1}); 
(2) next, press the start button to start a conversation (Figure \ref{fig:screenshot_2}).
At each turn, the facial expression of the virtual assistant would change along with the emotion of the generated utterance (Figure \ref{fig:screenshot_3}).
\textbf{The demo video is available at \url{https://youtu.be/Qy9mvyfc8eQ}}.

% celebA1948
% RaFD535

% NLG:
% --epoch 8
% --lr 1e-6 
% --batch_size 2 
% total loss是1e-7*lm_loss + emotion_cls_loss

% Face:
% 1300 epoch
% image size 256 x 256
% cond_nc (action vector的維度）35
% batch size 8
% Discriminator & Generator的lr都是1e-4

\subsection{Training Details}
In the experiments, we train the language generator by \textit{Adam} optimizer with each batch of 2 examples, 8 training epochs were performed without early stop.
The learning rate is 1e-6, and the weights for losses $\lambda_\text{u}$ and  $\lambda_\text{e} $ are 1e-7 and 1 respectively.
The entire implementation was based on PyTorch and HuggingFace transformers\footnote{\url{https://huggingface.co/transformers/}} package.
The medium size of DialoGPT is adopted as the basis of the model.

For the facial expression generator, we use \textit{Adam} as the optimizer with each batch of 8 examples, 1300 training epochs were performed without early stop.
The learning rates of discriminator and generator are both 1e-4.
The image size is $256 \times 256$, and the dimension of the action unit vector is 35. 
The other hyper-parameters and implementation details of the facial expression generator are same as the original work \cite{pumarola2018ganimation}. 

\begin{figure}[t!]
\centerline{\includegraphics[width=\linewidth]{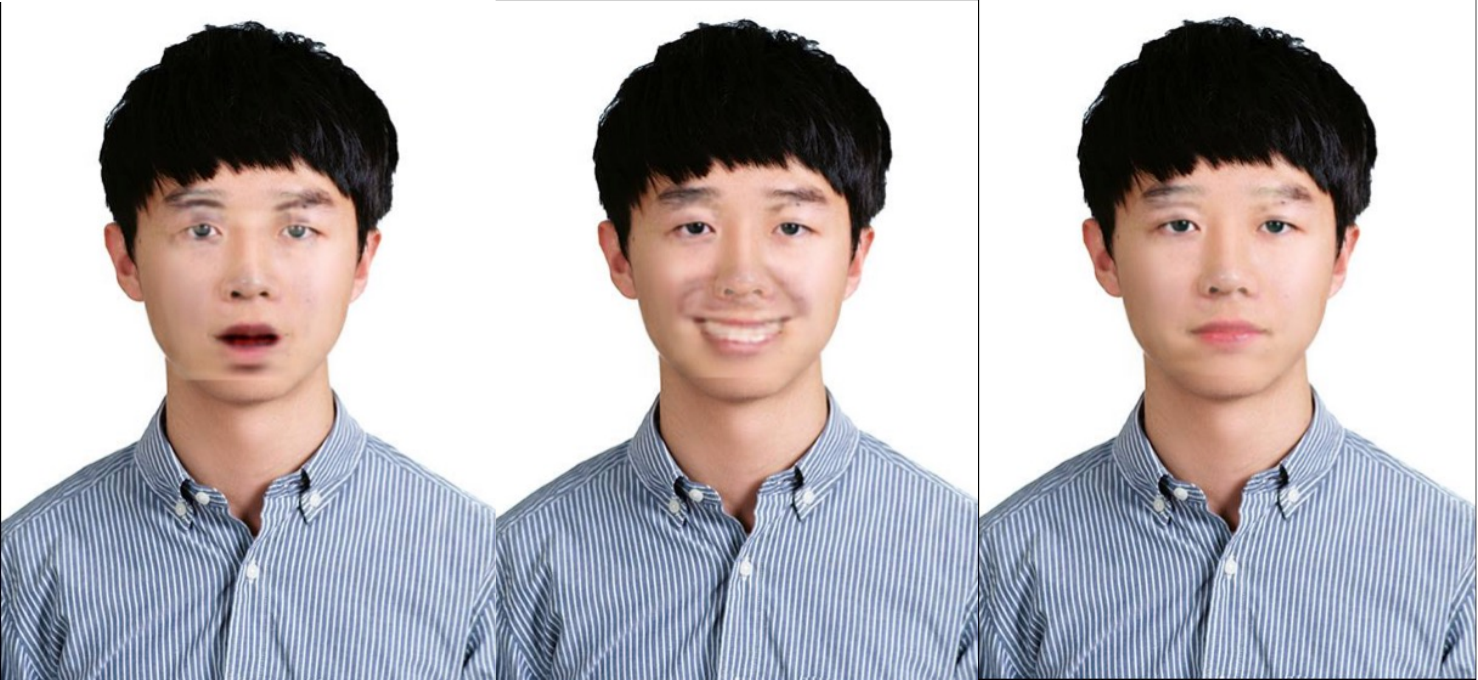}}
\vspace{-2mm}
	\caption{The generated results of the facial expression generator, the underlying emotions are surprise, happy, and contemptuous (left to right). }
	\label{fig:examples}
\end{figure}

% The generated example of our proposed pipeline is shown in Figure \ref{fig:example}, demonstrating the ability of generating sentences and facial expressions grounding to the same emotion signals.  

\subsection{Challenges Ahead}
HUMBO is a complex system consisting of multiple research problems, hence the data would be the key problem.
For example, one of the data sources of EmotionLines \cite{chen2018emotionlines} is also the TV show \emph{Friends}, but the emotions are labeled according to Ekman’s basic emotions \cite{ekman1997universal}, which is slightly different to the dataset we conduct \cite{zahiri2017emotion}.
The reason why we discard EmotionLines is because it has severe data-imbalance issue.

\begin{figure}[t!]
\centerline{\includegraphics[width=\linewidth]{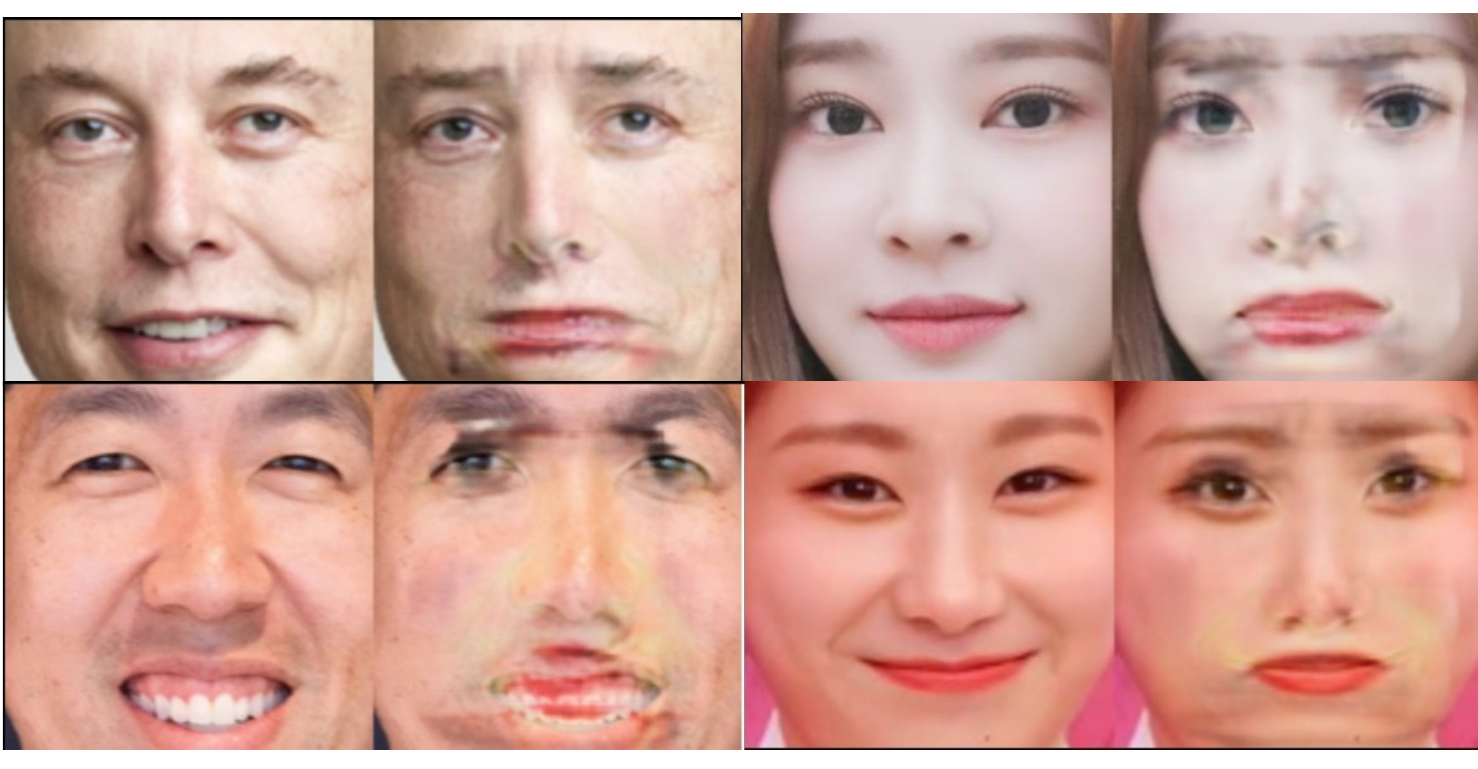}}
\vspace{-2mm}
	\caption{The failure cases of the generated results, the conditioned emotions are all "sad" (the left ones are the original images). }
	\label{fig:examples_2}
\end{figure}

In language part, we could imagine that there might be some easy methods to improve the performance, for example, conducting a more powerful and public pre-trained language model like GPT-3 \cite{brown2020language}.  
However, it would be much more difficult to improve the facial expression generator. 
Facial expression generation is the main highlighted feature of HUMBO, the goal is to endure users with the ability of determining the appearances of the virtual assistants by a single image.
Figure \ref{fig:examples} shows different generated faces of a person, we could observe that though not being perfect, the overall results are acceptable.
There are some obvious defects like unexpected shading and wrinkle, especially around the lips and eyebrows, we speculate that it is because those areas contain many Action Units.
The characteristics of original user-uploaded images are also important, Figure \ref{fig:examples_2} shows some failure cases, where we could observe that skin colors, wrinkles, and original facial expressions are critical.

\section{Conclusions}
In this work, we present HUMBO, a system aiming at generating dialogue responses and  simultaneously synthesize corresponding visual expressions on faces for better multimodal interaction.
% HUMBO can 
% (1) let users to determine the appearances of virtual assistants by a single image, and (2) generate coherent emotional utterances and facial expressions on the user-provided image.
HUMBO can let users determine the appearances of virtual assistants by a single image, and generate coherent emotional utterances and facial expressions on the user-provided image.
To bridge these two problem, we further propose to model the shared semantics of
the two modalities by emotion signals.
This is not only a brand new research direction but more importantly, an ultimate step toward more human-like virtual assistants.
% HUMBO can let users to determine the appearances of virtual assistants by a single image, and generate coherent emotional utterances and facial expressions on the user-provided image.

% In this paper, we introduce a new multimodal task which aims to generate natural language sentences and simultaneously synthesize corresponding facial expression. % according to the given attributes.  
% To bridge these two problem, we further propose to model the shared semantics of
% the two modalities by emotion signals.
% Furthermore, a dataset preparation strategy and a training framework are also introduced.
% % Since emotions are expressed through a combination of verbal and non-verbal channels, like gestures, facial expression, speech, and spoken content, therefore it could be viewed as the intersection of various modes.
% The proposed concept is a new research direction of multimodal dialogues and also an
% ultimate step towards more human-like interfaces of virtual assistants.
% Furthermore, a dataset preparation strategy and a training framework are also introduced.

\bibliography{anthology,custom}

\begin{thebibliography}{21}
\expandafter\ifx\csname natexlab\endcsname\relax\def\natexlab#1{#1}\fi

\bibitem[{Baltrusaitis et~al.(2018)Baltrusaitis, Zadeh, Lim, and
  Morency}]{baltrusaitis2018openface}
Tadas Baltrusaitis, Amir Zadeh, Yao~Chong Lim, and Louis-Philippe Morency.
  2018.
\newblock Openface 2.0: Facial behavior analysis toolkit.
\newblock In \emph{2018 13th IEEE International Conference on Automatic Face \&
  Gesture Recognition (FG 2018)}, pages 59--66. IEEE.

\bibitem[{Bordes et~al.(2017)Bordes, Boureau, and Weston}]{bordes2017learning}
Antoine Bordes, Y-Lan Boureau, and Jason Weston. 2017.
\newblock Learning end-to-end goal-oriented dialog.
\newblock In \emph{Proceedings of ICLR}.

\bibitem[{Brown et~al.(2020)Brown, Mann, Ryder, Subbiah, Kaplan, Dhariwal,
  Neelakantan, Shyam, Sastry, Askell, Agarwal, Herbert-Voss, Krueger, Henighan,
  Child, Ramesh, Ziegler, Wu, Winter, Hesse, Chen, Sigler, Litwin, Gray, Chess,
  Clark, Berner, McCandlish, Radford, Sutskever, and
  Amodei}]{brown2020language}
Tom~B. Brown, Benjamin Mann, Nick Ryder, Melanie Subbiah, Jared Kaplan,
  Prafulla Dhariwal, Arvind Neelakantan, Pranav Shyam, Girish Sastry, Amanda
  Askell, Sandhini Agarwal, Ariel Herbert-Voss, Gretchen Krueger, Tom Henighan,
  Rewon Child, Aditya Ramesh, Daniel~M. Ziegler, Jeffrey Wu, Clemens Winter,
  Christopher Hesse, Mark Chen, Eric Sigler, Mateusz Litwin, Scott Gray,
  Benjamin Chess, Jack Clark, Christopher Berner, Sam McCandlish, Alec Radford,
  Ilya Sutskever, and Dario Amodei. 2020.
\newblock \href {http://arxiv.org/abs/2005.14165} {Language models are few-shot
  learners}.

\bibitem[{Busso et~al.(2008)Busso, Bulut, Lee, Kazemzadeh, Mower, Kim, Chang,
  Lee, and Narayanan}]{busso2008iemocap}
Carlos Busso, Murtaza Bulut, Chi-Chun Lee, Abe Kazemzadeh, Emily Mower, Samuel
  Kim, Jeannette~N Chang, Sungbok Lee, and Shrikanth~S Narayanan. 2008.
\newblock Iemocap: Interactive emotional dyadic motion capture database.
\newblock \emph{Language resources and evaluation}, 42(4):335.

\bibitem[{Chen et~al.(2018)Chen, Hsu, Kuo, Ku et~al.}]{chen2018emotionlines}
Sheng-Yeh Chen, Chao-Chun Hsu, Chuan-Chun Kuo, Lun-Wei Ku, et~al. 2018.
\newblock Emotionlines: An emotion corpus of multi-party conversations.
\newblock \emph{arXiv preprint arXiv:1802.08379}.

\bibitem[{Ekman and Friesen(1976)}]{ekman1976measuring}
Paul Ekman and Wallace~V Friesen. 1976.
\newblock Measuring facial movement.
\newblock \emph{Environmental psychology and nonverbal behavior}, 1(1):56--75.

\bibitem[{Ekman and Keltner(1997)}]{ekman1997universal}
Paul Ekman and Dacher Keltner. 1997.
\newblock Universal facial expressions of emotion.
\newblock \emph{Segerstrale U, P. Molnar P, eds. Nonverbal communication: Where
  nature meets culture}, 27:46.

\bibitem[{Gulrajani et~al.(2017)Gulrajani, Ahmed, Arjovsky, Dumoulin, and
  Courville}]{gulrajani2017improved}
Ishaan Gulrajani, Faruk Ahmed, Martin Arjovsky, Vincent Dumoulin, and Aaron~C
  Courville. 2017.
\newblock Improved training of wasserstein gans.
\newblock In \emph{Advances in Neural Information Processing Systems}, pages
  5767--5777.

\bibitem[{Hakkani-T{\"u}r et~al.(2016)Hakkani-T{\"u}r, T{\"u}r, Celikyilmaz,
  Chen, Gao, Deng, and Wang}]{hakkani2016multi}
Dilek Hakkani-T{\"u}r, G{\"o}khan T{\"u}r, Asli Celikyilmaz, Yun-Nung Chen,
  Jianfeng Gao, Li~Deng, and Ye-Yi Wang. 2016.
\newblock Multi-domain joint semantic frame parsing using bi-directional
  rnn-lstm.
\newblock In \emph{Proceedings of INTERSPEECH}, pages 715--719.

\bibitem[{Isola et~al.(2017)Isola, Zhu, Zhou, and Efros}]{isola2017image}
Phillip Isola, Jun-Yan Zhu, Tinghui Zhou, and Alexei~A Efros. 2017.
\newblock Image-to-image translation with conditional adversarial networks.
\newblock In \emph{Proceedings of the IEEE conference on computer vision and
  pattern recognition}, pages 1125--1134.

\bibitem[{Langner et~al.(2010)Langner, Dotsch, Bijlstra, Wigboldus, Hawk, and
  Van~Knippenberg}]{langner2010presentation}
Oliver Langner, Ron Dotsch, Gijsbert Bijlstra, Daniel~HJ Wigboldus, Skyler~T
  Hawk, and AD~Van~Knippenberg. 2010.
\newblock Presentation and validation of the radboud faces database.
\newblock \emph{Cognition and emotion}, 24(8):1377--1388.

\bibitem[{Liu et~al.(2015)Liu, Luo, Wang, and Tang}]{liu2015faceattributes}
Ziwei Liu, Ping Luo, Xiaogang Wang, and Xiaoou Tang. 2015.
\newblock Deep learning face attributes in the wild.
\newblock In \emph{Proceedings of International Conference on Computer Vision
  (ICCV)}.

\bibitem[{Pumarola et~al.(2018)Pumarola, Agudo, Martinez, Sanfeliu, and
  Moreno-Noguer}]{pumarola2018ganimation}
Albert Pumarola, Antonio Agudo, Aleix~M Martinez, Alberto Sanfeliu, and
  Francesc Moreno-Noguer. 2018.
\newblock Ganimation: Anatomically-aware facial animation from a single image.
\newblock In \emph{Proceedings of the European Conference on Computer Vision
  (ECCV)}, pages 818--833.

\bibitem[{Radford et~al.(2019)Radford, Wu, Child, Luan, Amodei, Sutskever
  et~al.}]{radford2019language}
Alec Radford, Jeffrey Wu, Rewon Child, David Luan, Dario Amodei, Ilya
  Sutskever, et~al. 2019.
\newblock Language models are unsupervised multitask learners.
\newblock \emph{OpenAI blog}, 1(8):9.

\bibitem[{Su et~al.(2018)Su, Li, Gao, Liu, and Chen}]{su2018discriminative}
Shang-Yu Su, Xiujun Li, Jianfeng Gao, Jingjing Liu, and Yun-Nung Chen. 2018.
\newblock Discriminative deep dyna-q: Robust planning for dialogue policy
  learning.
\newblock In \emph{Proceedings of the 2018 Conference on Empirical Methods in
  Natural Language Processing}, pages 3813--3823.

\bibitem[{Suwajanakorn et~al.(2017)Suwajanakorn, Seitz, and
  Kemelmacher-Shlizerman}]{suwajanakorn2017synthesizing}
Supasorn Suwajanakorn, Steven~M Seitz, and Ira Kemelmacher-Shlizerman. 2017.
\newblock Synthesizing obama: learning lip sync from audio.
\newblock \emph{ACM Transactions on Graphics (TOG)}, 36(4):95.

\bibitem[{Wen et~al.(2017)Wen, Gasic, Mrksic, Rojas-Barahona, Su, Ultes,
  Vandyke, and Young}]{wen2017network}
Tsung-Hsien Wen, Milica Gasic, Nikola Mrksic, Lina~M Rojas-Barahona, Pei-Hao
  Su, Stefan Ultes, David Vandyke, and Steve Young. 2017.
\newblock A network-based end-to-end trainable task-oriented dialogue system.
\newblock In \emph{Proceedings of EACL}, pages 438--449.

\bibitem[{Willcox(1982)}]{Willcox1982TheFW}
G.~Willcox. 1982.
\newblock The feeling wheel a tool for expanding awareness of emotions and
  increasing spontaneity and intimacy.
\newblock \emph{Transactional Analysis Journal}, 12:274--276.

\bibitem[{Zahiri and Choi(2017)}]{zahiri2017emotion}
Sayyed~M. Zahiri and Jinho~D. Choi. 2017.
\newblock \href {http://arxiv.org/abs/1708.04299} {Emotion detection on tv show
  transcripts with sequence-based convolutional neural networks}.

\bibitem[{Zhang et~al.(2020)Zhang, Sun, Galley, Chen, Brockett, Gao, Gao, Liu,
  and Dolan}]{Zhang_2020}
Yizhe Zhang, Siqi Sun, Michel Galley, Yen-Chun Chen, Chris Brockett, Xiang Gao,
  Jianfeng Gao, Jingjing Liu, and Bill Dolan. 2020.
\newblock \href {https://doi.org/10.18653/v1/2020.acl-demos.30} {Dialogpt :
  Large-scale generative pre-training for conversational response generation}.
\newblock \emph{Proceedings of the 58th Annual Meeting of the Association for
  Computational Linguistics: System Demonstrations}.

\bibitem[{Zhu et~al.(2017)Zhu, Park, Isola, and Efros}]{zhu2017unpaired}
Jun-Yan Zhu, Taesung Park, Phillip Isola, and Alexei~A Efros. 2017.
\newblock Unpaired image-to-image translation using cycle-consistent
  adversarial networks.
\newblock In \emph{Proceedings of the IEEE International Conference on Computer
  Vision}, pages 2223--2232.

\end{thebibliography}
\bibliographystyle{acl_natbib}

% \appendix

% \section{Example Appendix}
% \label{sec:appendix}

% This is an appendix.

\end{document}